\documentclass[11pt]{article}

\usepackage[final]{acl}

\usepackage{times}
\usepackage{latexsym}
\usepackage[most]{tcolorbox}

\usepackage[T1]{fontenc}
\usepackage[utf8]{inputenc}
\usepackage{hyperref}

\usepackage{microtype}

\usepackage{inconsolata}

\usepackage{graphicx}

\usepackage{fvextra} %
\DefineVerbatimEnvironment{PromptVerbatim}{Verbatim}{
  fontsize=\small,
  breaklines=true,
  breakanywhere=true,
  commandchars=\\\{\}
}
\newtcolorbox{promptbox}[1]{
  enhanced,
  breakable,
  colback=white,
  colframe=black,
  boxrule=0.5pt,
  arc=2pt,
  left=6pt,right=6pt,top=6pt,bottom=6pt,
  title=\textbf{#1},
  fonttitle=\small,
}

\newcommand\blfootnote[1]{
    \begingroup
    \renewcommand\thefootnote{}\footnote{#1}
    \addtocounter{footnote}{-1}
    \endgroup
}

\title{RETUYT-INCO at BEA 2026 Shared Task 2:\\ Meta-prompting in Rubric-based Scoring for German}

\author{
  Ignacio Sastre$^\dagger$ \and Ignacio Remersaro \\
  \bf Facundo Díaz \and Nicolás de Horta \\
  \bf Luis Chiruzzo \and Aiala Rosá \and Santiago Góngora \\
  Instituto de Computación, Facultad de Ingeniería, Universidad de la República \\
  Montevideo, Uruguay
}

\begin{document}
\maketitle
\begin{abstract}
In this paper, we present the RETUYT-INCO participation at the BEA 2026 shared task ``Rubric-based Short Answer Scoring for German''.
Our team participated in track 1 (\textit{Unseen answers three-way}), track 3 (\textit{Unseen answers two-way}) and track 4 (\textit{Unseen questions two-way}).
Since these tracks required scoring short student answers using specific rubrics, we looked for ways to handle the changing nature of the task. 
We created a method called \textbf{Meta-prompting}. 
In this approach, an LLM creates a custom prompt based on examples from the \texttt{Train} set. 
This prompt is then used to grade new student answers. 
Along with this method, we also describe other approaches we used, such as classic machine learning, fine-tuning open-source LLMs, and different prompting techniques.
According to the official results, our team placed 6th out of 8 participants in Track 1 with a QWK of 0.729. 
In Track 3, we secured 4th place out of 9 with a QWK of 0.674, and we also placed 4th out of 8 in Track 4 with a QWK of 0.49.

\end{abstract}

\section{Introduction}

\blfootnote{$\dagger$ Corresponding author: \texttt{isastre@fing.edu.uy}.}

Short-answer scoring is a promising research direction in the field of Natural Language Processing for education. 
Unlike multiple-choice tests, which only require students to recognize a correct answer, short answers require them to actively construct a response using the concepts they have learned from the curriculum. 
This may be appropriate for capturing a student's true understanding of a subject.
The automation of short-text assessment has the potential to deliver faster feedback and consistent grading, offering vital support to overburdened educators, especially in under-resourced environments. 
Over the years, researchers have used various benchmarks and shared tasks to refine these systems~\cite{burrows_eras_2015,bai_survey_2023,dzikovska-etal-2013-semeval}.
We, as a research group, are deeply interested in these applications of NLP tools~\cite{chiruzzo-etal-2022-using,rosa2025platform}.

In this paper, we present the RETUYT-INCO participation in the BEA 2026 Shared Task 2: Rubric-based Short Answer Scoring for German \cite{gombert2026bea_shared_task}. 
This task involves the automatic evaluation of short-answer responses in the domains of science and mathematics, framed as a classification problem with both two-way and three-way scoring rubrics.
Our team participated in Tracks 1 (Unseen answers three-way), 3 (Unseen answers two-way) and 4 (Unseen questions two-way).
Our team has participated in various shared tasks of the BEA workshop in recent years~\cite{baladon-etal-2023-retuyt,sastre-etal-2024-retuyt,gongora-etal-2025-retuyt}, generally exploring solutions that do not require extensive computational resources. 
Additionally, this year we were able to incorporate state-of-the-art Gemini models into our experiments through the \textit{Gemini Academic Program}.

\section{Meta-prompting}

\begin{figure*}
    \centering
    \includegraphics[width=0.72\linewidth]{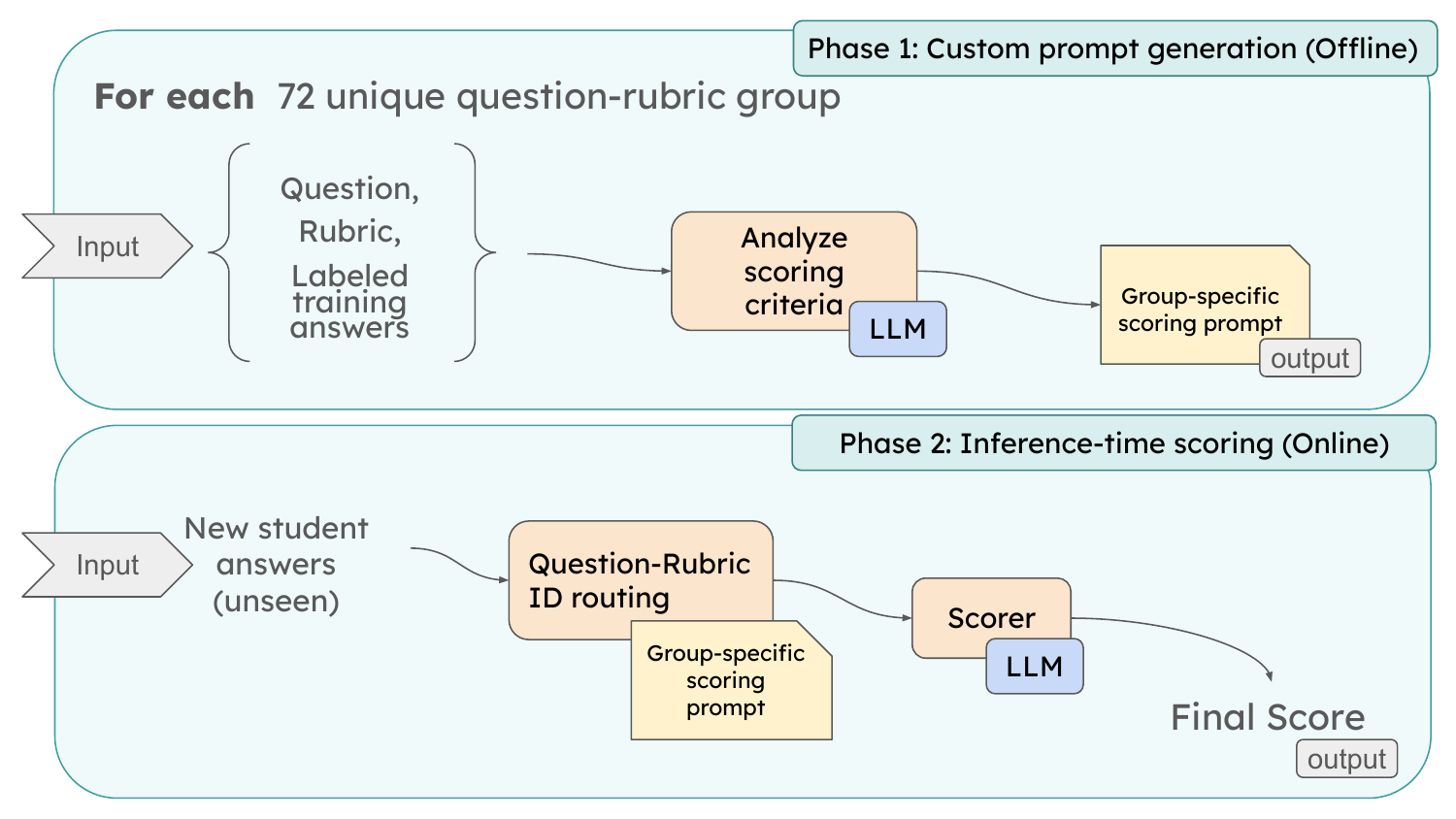}
    \caption{Meta-prompting operation scheme in its two phases}
    \label{fig:metaprompting}
\end{figure*}

We mainly focus on the \textit{unseen answer} tracks, in which systems must score new student answers for question-rubric pairs that have already been observed in the training data. An analysis of the \texttt{Train} split revealed 72 distinct question-rubric groups, each defined by a fixed question and its associated rubric. As a baseline, we used Gemini 3 Flash\footnote{\url{https://blog.google/products-and-platforms/products/gemini/gemini-3-flash/}} with a generic scoring prompt instantiated with the corresponding question and rubric. On the \texttt{Trial} set, this baseline exhibited substantial performance variation across groups, suggesting that a single generic prompt does not capture the scoring criteria equally well for all question-rubric pairs.

To address the performance inconsistencies observed with generic scoring prompts, we leverage established advances in automated prompt optimization. This field has progressed from Instruction Induction \cite{honovich2022instructioninductionexamplesnatural}, which identifies task descriptions from few-shot examples, to the Automatic Prompt Engineer \cite{zhou2023large} framework for systematic search and scoring. Most recently, GEPA \cite{gepa} introduces a reflexive evolution approach that uses genetic algorithms and natural language reasoning to iteratively refine instructions, offering a robust alternative to static baselines for the diverse requirements of rubric-based scoring.

Based on this line of work, we introduce the \textbf{meta-prompting} approach, that induces a custom scoring prompt for each question-rubric group. The method consists of two phases. In the first phase, for each group, we prompt an LLM with the question, the rubric, and the full set of labeled training answers for that group, serialized as JSON. The model is asked to generate a reusable scoring prompt tailored to that specific group. In the second phase, at inference time, each new student answer is routed to its corresponding question-rubric group and evaluated using the group-specific prompt generated in the first phase. 
Figure~\ref{fig:metaprompting} shows a diagram of the method.

To further improve performance, we generate multiple candidate prompts for each question-rubric group, varying the prompt formulation (see Appendix~\ref{sec:appendix_metaprompts}), the thinking budget (medium or high), and the use of synthetic examples (explained in Section~\ref{sec:synthetic-data}). For each group, we then retain the prompt that performs best on the \texttt{Trial} set. Table~\ref{tab:trial-results} reports all evaluated variants and their corresponding weighted F1 and QWK scores.

\subsection{Meta-prompt generation}

For each question-rubric group, we generate a dedicated scoring prompt in an offline step using the fixed question, its rubric, and the labeled training answers associated with that group. The training instances are provided as a JSON object containing each answer and its gold label. An LLM then infers the scoring criteria that best characterize the group and produces a reusable prompt template for evaluating unseen answers.
The meta-prompt encourages the model to capture the semantic distinctions underlying the rubric instead of relying on superficial lexical patterns. The resulting template includes placeholders for the \textit{question}, \textit{rubric}, and \textit{answer}, and instructs the downstream model to output only the final \textit{label}. This process yields one candidate scoring prompt per group, which can later be used directly or compared with alternative prompt variants. The complete prompt template is provided in Appendix~\ref{sec:appendix_baseline_prompt}.

\begin{table*}[ht!]
\centering
\begin{tabular}{lcc}
\hline
\textbf{Approach} & \textbf{Weighted F1} & \textbf{QWK} \\
\hline
\multicolumn{3}{l}{\textbf{Baselines}} \\
\hline
Prompting Llama 3.1 8B Instruct & 0.610 & 0.050 \\
Prompting Gemini 3 Flash & 0.820 & 0.563 \\
\hline
\multicolumn{3}{l}{\textbf{Classical Machine Learning}} \\
\hline
SVM + TF-IDF (2-10)grams & 0.815 & 0.545 \\
\hline
\multicolumn{3}{l}{\textbf{Tuning Llama 3.1 8B}} \\
\hline
Fine-tuning LoRA & 0.808 & 0.523 \\
Prompt tuning & 0.824 & 0.577 \\
Prompt tuning with synthetic data & 0.826 & 0.583 \\
Ensemble prompt tuning with Gemini baseline & 0.832 & 0.595 \\
\hline
\multicolumn{3}{l}{\textbf{Meta-prompting}} \\
\hline
Meta-prompt thinking medium & 0.851 & 0.649 \\
Meta-prompt thinking high & 0.845 & 0.635 \\
Meta-prompt thinking high with synthetic data & 0.856 & 0.659 \\
Meta-prompt thinking high (different prompt) & 0.859 & 0.661 \\
Meta-prompt best variant per group & \textbf{0.892} & \textbf{0.743} \\
\hline
\end{tabular}
\caption{Results on the \texttt{Trial} set. In bold, the best result for each metric.}
\label{tab:trial-results}
\end{table*}

\subsection{Synthetic data}
\label{sec:synthetic-data}

As part of a strategy to normalize the data imbalance within each question, we implemented a synthetic data generation module.

The pipeline is implemented as a state graph (using LangGraph\footnote{\url{https://github.com/langchain-ai/langgraph}} and LangChain\footnote{\url{https://github.com/langchain-ai/langchain}}).
Given a question, a student answer, and the rubric description only for the target label, the LLM is prompted to produce a new student answer matching that quality level. 
The prompt instructs the model to role-play as a real student, explicitly forbidding meta-commentary, references to grading criteria, or verbatim copying of the reference answer. 
Additionally, all previously generated answers for the same question are included in the prompt, instructing the model to produce a substantially different response in terms of wording, structure, and reasoning.

This is followed by a noise injection module to add grammatical errors and colloquialisms that mimic real-world writing.
Finally, each response is validated against the full rubric and a conditional retry system that discards inconsistent examples to ensure the quality of the synthetic dataset.

We applied this method to the 25 questions with the greatest imbalance between correct and incorrect answers in the \texttt{Train} set, producing the number of samples necessary to reduce the gap between the two categories by half.

\section{Other approaches}

Apart from meta-prompting, we explored other approaches ranging from classic machine learning algorithms to fine-tuning open LLMs.

\subsection{Role-playing prompting}
\label{sec:roleplaying}

Vaguely inspired in contemporary works that report Role-playing prompting may change the expected behavior of an LLM~\cite{roleplay_jailbreaking_1, wang-etal-2025-delman, luo-etal-2026-simple}, we wanted to give a try on a prompt that described the classification task differently.

Therefore, we prompted Gemini 3 flash to act as isolated teachers in a \textit{distant planet}.
The API was called three times, always giving as an input the original question, the student's answer, and the rubric. 
The first call prompts the model to act as a teacher that focus on the positive side of the student's answer, while the second one focuses on the negative aspects.
The outputs of those calls are two reviews given as an input to the third teacher (a new API call), acting as a meta-reviewer that has to make the final decision.
Given how expensive it was to run this experiment (it takes 3 API calls to classify each instance), we only submitted its predictions for the \textit{2 way unseen questions} track, without evaluating on the \texttt{Trial} set.
Appendix ~\ref{sec:appendix_roleplaying_prompts} includes the prompts used for this approach.

\begin{table*}[ht!]
\centering
\begin{tabular}{lccc}
\hline
\textbf{Submission} & \textbf{QWK} & \textbf{WF1} & \textbf{\#} \\
\hline
\multicolumn{4}{c}{\textbf{Track 1 - Unseen Answers 3-Way}} \\
\hline
Meta-prompt thinking medium & \textbf{0.729} & \textbf{0.728} & \textbf{30/44} \\
Meta-prompt thinking high (different prompt) & 0.696 & 0.702 & 35/44\\
\hline
\multicolumn{4}{c}{\textbf{Track 3 - Unseen Answers 2-Way}} \\
\hline
Baseline prompting Gemini 3 Flash & 0.598 & 0.835 & 37/51 \\
SVM + TF-IDF (2-10)grams & 0.520 & 0.806 & 43/51 \\
Prompt tuning with synthetic data & 0.492 & 0.786 & 44/51 \\
Ensemble prompt tuning with Gemini baseline  & 0.537 & 0.809 & 42/51 \\
Meta-prompt thinking medium & 0.654 & 0.853 & 25/51 \\
Meta-prompt best variant per group & \textbf{0.674} & \textbf{0.863} & \textbf{14/51} \\
\hline
\multicolumn{4}{c}{\textbf{Track 4 - Unseen Questions 2-Way}} \\
\hline
Baseline prompting Gemini 3 Flash & \textbf{0.490} & \textbf{0.789} & \textbf{12/39} \\
SVM + TF-IDF (2-10)grams & 0.341 & 0.736 & 38/39 \\
Roleplaying Gemini 3 Flash & 0.432 & 0.766 & 28/39 \\
\hline
\end{tabular}
\caption{Results for the three tracks evaluated. The QWK metric is used as the primary reference. The "\#" column indicates the ranking position out of all submissions.}
\label{tab:submissions}
\end{table*}

\subsection{Prompt tuning and LoRA fine-tuning}

Following our approach in previous shared tasks, where we fine-tuned small LLMs~\cite{baladon-etal-2023-retuyt,sastre-etal-2024-retuyt,gongora-etal-2025-retuyt}, we also explored this strategy using Llama 3.1 8B Instruct~\cite{grattafiori2024llama3herdmodels}.

In the unseen answers tracks, the task can be naturally decomposed into 72 question-rubric groups, each associated with its own subset of training examples. A straightforward approach would therefore be to fine-tune one model per group. However, performing 72 separate fine-tuning runs, even with a parameter-efficient method such as Low-Rank Adaptation (LoRA)~\cite{hu2022lora}, would be computationally expensive and time-consuming. In addition, the amount of training data available for each group is often limited and may be highly imbalanced.

For this reason, we explored Prompt Tuning~\cite{lester-etal-2021-power}, in which the base model is kept frozen and only a small set of input embeddings, known as a soft prompt, is learned. Recent work on Concept Tokens~\cite{concept_token} has shown that, in Llama 3.1 8B, learning a single embedding using a definitional corpus can be sufficient to steer model behavior. Inspired by this idea, we learned one embedding per question-rubric group. Each embedding was trained using the corresponding training examples, augmented with synthetic data (see Section~\ref{sec:synthetic-data}), while the \texttt{Trial} split was used for early stopping. At inference time, given a new answer, we construct a prompt containing the question together with the learned embedding associated with its question-rubric group.

We compared this method against a single LoRA fine-tuning run over the full \texttt{Train} set. As shown in Table~\ref{tab:trial-results}, prompt tuning achieved better performance on the \texttt{Trial} set than this global LoRA baseline. Both adaptation methods substantially outperformed prompting alone with the same small open-source model, which obtained a QWK of 0.050, whereas prompt tuning reached 0.583. This also highlights the large gap between prompting small open-source LLMs and prompting strong proprietary models: under the same general prompting setup, Gemini 3 Flash achieved a QWK of 0.56.

\subsection{Support Vector Machines}

Since this was ultimately a classification task, we wanted to attempt it using at least one classic machine learning algorithm. 
We began with the premise that the scores might be inferable from the student's writing alone, without the need for additional context.
While we anticipated that this model would not compete with neural architectures, following our previous work in BEA shared tasks~\cite{sastre-etal-2024-retuyt,gongora-etal-2025-retuyt}, we consider it worthwhile to quantify the widening performance gap relative to modern transformer-based models.

Using Scikit-learn~\cite{pedregosa2011scikit}, we conducted several preliminary experiments. 
We tested a wide range of statistical classifiers—including Naive Bayes, Random Forest, and Support Vector Machines—alongside various bag-of-words configurations for text representation at both word and character levels.
The best configuration we found when evaluating on the \texttt{Trial} set was feeding a sigmoid-kernel SVC\footnote{\url{https://scikit-learn.org/stable/modules/generated/sklearn.svm.SVC.html}}, with the \texttt{Train} set represented by a TF-iDF vectorizer\footnote{\url{https://scikit-learn.org/stable/modules/generated/sklearn.feature\_extraction.text.TfidfVectorizer.html}} using character-level n-grams in a $[2,10]$ range.
We trained and submitted a model using this configuration for the \textit{2-way unseen answers} track and another one for the \textit{2-way unseen questions} track.

\section{Results}

Table~\ref{tab:submissions} summarizes the performance of our submitted systems across the three shared-task tracks in which we participated.
Our main focus was Track~3 (two-way unseen answers), for which we submitted the Gemini 3 Flash prompting baseline, the SVM model, the prompt tuning approach based on Llama~3.1~8B, an ensemble of the baseline and prompt tuning methods, and several variants of the meta-prompting technique. The ensemble combines the baseline and prompt-tuning systems by selecting, for each question--rubric group, the method that achieved the best performance on the \texttt{Trial} set.

For Track~4 (two-way unseen questions), where the questions and rubrics are not available in advance, we submitted the same Gemini 3 Flash baseline, the SVM model, and the role-playing method. The remaining approaches depend on prior access to the question--rubric groups and therefore are not applicable in this setting. Finally, for Track~1 (three-way unseen answers), we submitted only variants of the meta-prompting approach, as this method yielded the strongest results on Track~3.

The only method that improved upon the prompting baseline was the meta-prompting approach. This result is likely due to the fact that the baseline relies on Gemini 3 Flash, a much larger and more recent model than the open model used in other experiments, Llama 3.1 8B. The strength of this baseline is further supported by its ranking in the top third of submissions on Track~4.

Meta-prompting and the role-playing strategy are the only other methods that use the same base LLM, making them the most directly comparable to the baseline. Among the Gemini 3 Flash-based approaches, meta-prompting achieved the strongest performance, improving over the baseline by nearly 8 points on QWK and ranking 14th out of 51 submissions overall. At the team level, this placed us 4th out of 9 teams that submitted to Track 3.

For Track~4, the baseline was our strongest submission, outperforming the role-playing strategy and placing us 4th out of 8 teams. Finally, we placed 6th out of 8 teams on Track~1 with the meta-prompting submission. It is worth noting that we did not tune the method on the \texttt{Trial} set for the three-way setting, and instead reused the same meta-prompting configuration developed for Track~3, which may partly explain the lower performance on this track.

Regarding the Llama 3.1 8B prompt tuning method, we observe a substantial drop in performance from the \texttt{Trial} set to the final \texttt{Test} set. A likely explanation is that the \texttt{Trial} set was used for early stopping, which may have caused the learned embeddings to overfit to those examples. This issue is further compounded by the fact that many question-rubric groups are both small and highly imbalanced, with some groups containing examples from only a single class in the \texttt{Trial} split.

\section{Conclusions}

In this paper, we presented the RETUYT-INCO participation in the BEA 2026 shared task, ``Rubric-based Short Answer Scoring for German''. 
Our team tried different approaches across Track 1 (Unseen answers 3-way), Track 3 (Unseen answers 2-way), and Track 4 (Unseen questions 2-way).

A central part of our experimentation involved \textbf{meta-prompting}, where an LLM generated scoring prompts based on available training instances; these prompts were then used as input for a subsequent LLM inference to classify unseen texts.
Alongside this technique, we evaluated the performance of traditional machine learning models and the fine-tuning of open-source LLMs.

The official results placed our systems in the mid-table across the three tracks.
In Track 1, we obtained a Quadratic Weighted Kappa (QWK) of $0.729$, placing 6th out of 8 participants. 
Our performance in the two-way tracks was notably robust, securing 4th place out of 9 in Track 3 with a QWK of $0.674$, and 4th place out of 8 in Track 4 with a QWK of $0.490$.

Future work should first explore how to better adapt the meta-prompting strategy to the three-way setting, where our submitted results were still preliminary.
Beyond this shared task, the approach should also be evaluated in other rubric-based scoring datasets, as well as in related educational NLP tasks, to assess its generalization and robustness.
Finally, our experiments highlight the substantial gap between large proprietary models and smaller open-source LLMs. Narrowing this gap remains an important direction, both to reduce dependency on closed systems and to make high-performing educational NLP tools more accessible and cost-effective.

\section*{Limitations}

Many of our experiments rely on closed proprietary models that are accessible only through provider APIs. Although these models are relevant from a research perspective due to their strong performance, this dependence limits the accessibility and applicability of the proposed methods in real educational contexts. This is especially important in settings such as rural schools, where internet connectivity, privacy constraints, and cost considerations may restrict the use of external APIs. Moreover, the meta-prompting and role-playing approaches require multiple LLM calls and consume a large number of tokens, increasing both cost and latency.

Our experiments with Llama~3.1~8B were conducted using a single Google Colab Pro subscription, which limited our ability to tune hyperparameters and perform ablation studies. Consequently, these results should be viewed as preliminary rather than fully optimized.
Regarding the generated synthetic data, we did not conduct a human evaluation of whether the generated samples were of good quality. 
We focused on the empirical side of this method, hence prioritizing seeing an improvement in the results. 
Additionally, we did not evaluate how the \textit{distant planet} setting in the role-playing approach impacted the performance of the method.

Finally, we explored the three-way setting less extensively than the two-way configurations. 
The submitted systems for this track were mostly adaptations of methods developed for the two-way unseen answers setting, rather than approaches specifically optimized for three-way scoring. 
Given the stronger results obtained in the two-way setting after task-specific tuning, we believe that better-adapted versions of our methods could substantially improve performance in the three-way setting.

\section*{Ethics}

Our research uses state-of-the-art, closed-source models. These models are proprietary and operate via external APIs, so they may not be suitable for processing sensitive data, particularly involving minors. 
The lack of transparency regarding data handling and privacy for minors presents a significant risk. 
We believe it should always be a priority, when using these solutions in real-world educational deployments, to use local, private infrastructure.

Furthermore, we acknowledge the substantial environmental impact associated with the deployment of these large-scale models. 
The energy consumption and resulting carbon footprint of state-of-the-art closed models are significant. 
Given that the work in this paper is a relatively constrained classification task, we emphasize that these state-of-the-art models may be computationally excessive. 
We encourage the community to explore smaller, specialized models that can provide a more sustainable and resource-efficient alternatives.

\section*{Acknowledgments}

This paper was funded by \textit{Agencia Nacional de Investigación e Innovación} (ANII, Uruguay), Project No. $FSED\_2\_2023\_1\_179355$.
Additionally,  this research is with support from Google.org and the Google Cloud Research Credits program for the Gemini Academic Program.

\bibliography{anthology,custom}

\appendix

\section{Baseline prompt}
\label{sec:appendix_baseline_prompt}

This Appendix presents the prompt used for the prompting baselines.

\begin{promptbox}{Baseline prompt}
\begin{PromptVerbatim}
You are an expert educational rater for rubric-based short-answer scoring.

Your task is to assign a binary score to a student's answer using the question and the rubric.

Label set for this 2-way setting:
- Output "Correct" only if the student's answer satisfies the rubric for "Correct".
- Output "Incorrect" otherwise.
- Answers that match "Partially Correct" must be labeled "Incorrect".

Decision procedure:
1. Read the question, the student answer, and all rubric levels.
2. Identify the full set of meaning requirements for the rubric label "Correct".
3. Use the question only as context to interpret the student's wording.
4. Do not use outside knowledge to add content that is not stated or clearly implied by the student's answer.
5. Accept paraphrases, synonyms, different wording, and short or fragmentary answers if their meaning clearly matches the rubric.
6. Output "Correct" only if all requirements for a fully correct answer are present and unambiguous.
7. Output "Incorrect" if any required element is missing, only partially present, too vague to verify, off-topic, self-contradictory, nonsensical, or incompatible with the rubric.
8. If the rubric allows multiple alternative ways to be fully correct, any one complete valid alternative is sufficient.
9. Ignore spelling and grammar errors unless they make the meaning unclear.
10. Ignore extra details unless they contradict the required content or make the answer incompatible with the rubric.
11. For multi-part requirements, all required parts must be present unless the rubric explicitly states otherwise.
12. Do not output any explanation.

The question, answer, and rubric may be in German. Score based on meaning, not language quality.

Input:
<Question>
\{question\}
</Question>

<StudentAnswer>
\{answer_to_classify\}
</StudentAnswer>

<Rubric>
<Incorrect>
\{rubric_incorrect\}
</Incorrect>
<PartiallyCorrect>
\{rubric_partially_correct\}
</PartiallyCorrect>
<Correct>
\{rubric_correct\}
</Correct>
</Rubric>

Return exactly one word and nothing else:
Correct
or
Incorrect
\end{PromptVerbatim}
\end{promptbox}

\section{Meta-prompting phase 1 prompts}
\label{sec:appendix_metaprompts}

This Appendix presents the prompts utilized in the Metaprompting approach, the second text is an extension from the metaprompt, that is appended before the return criteria.

\begin{promptbox}{Meta-prompt}
\begin{PromptVerbatim}
You are an expert in educational assessment and prompt design.

Given a question, a rubric, and labeled training answers for one fixed question/rubric group, generate a reusable prompt for scoring a NEW answer for that same group.
The possible labels are Correct and Incorrect (Partially Correct are classified as Incorrect).

The generated prompt must:
- be specific to this question/rubric group,
- reflect the rubric and the training signals,
- use training examples for illustrating the classification instructions,
- instruct the model to output only the final label and nothing else,
- contain exactly these placeholders:
  - \{question\}
  - \{answer\_to\_classify\}

Do not:
 - Repeat the rubric as-is: the idea is for you to reflect on the training examples and generate better criteria,
 - Add examples as few-shot: use the examples to explain the task and classification criteria, and to illustrate your explanations.

Return only the generated prompt template as plain text.
Do not include any explanation or extra text.

Input:
<Question>
\{question\}
</Question>

<Rubric>
<Incorrect>
\{rubric\_incorrect\}
</Incorrect>
<PartiallyCorrect>
\{rubric\_partially\_correct\}
</PartiallyCorrect>
<Correct>
\{rubric\_correct\}
</Correct>
</Rubric>

<TrainingExamples>
\{training\_examples\}
</TrainingExamples>
\end{PromptVerbatim}
\end{promptbox}

\begin{promptbox}{Meta-prompt extension}
\begin{PromptVerbatim}
The expected prompt should be sufficiently complex, nuanced, and comprehensive so that, if you were given only your own prompt, you would be fully confident in your ability to correctly classify all training examples, without overfitting to the specific examples provided. Do not optimize for brevity at the expense of completeness. It is perfectly acceptable for the prompt to be long if that is necessary to capture all relevant distinctions, edge cases, and decision criteria required for perfect classification. Ask yourself: does this prompt contain all the information needed to classify every example correctly?
\end{PromptVerbatim}
\end{promptbox}

\section{Roleplaying prompts}
\label{sec:appendix_roleplaying_prompts}

This Appendix presents the prompts used in the \textit{Role-playing prompting} approach, described in Section~\ref{sec:roleplaying}.

\begin{promptbox}{Positive reviewer}
\begin{PromptVerbatim}
You are a teacher in a distant planet correcting homework in german. You do not have the possibility of communicating with any of your colleagues.
Consider the following question in German written by some teacher in your solar system: '{question}'. A high-school student wrote this answer: '\{answer\}'.

Your task is to generate a three-paragraph discussion on reasons to consider that answer as Correct or Incorrect. In order to provide a framework, you can consider the answer is Correct or Incorrect according to the following criteria:
- The answer is Incorrect if: \{incorrect\}
- The answer is Correct if: \{correct\}

In your discussion try to focus on the POSITIVE aspects of the answer: why would you consider it as Correct? Write your review in English.
\end{PromptVerbatim}
\end{promptbox}

\begin{promptbox}{Negative reviewer}
\begin{PromptVerbatim}
You are a teacher in a distant planet correcting homeworks in german. You do not have the possibility of communicating with any of your colleagues.
Consider the following question in German written by some teacher in your solar system: '\{question\}'. A high-school student wrote this answer: '\{answer\}'.

Your task is to generate a three-paragraph discussion on reasons to consider that answer as Correct or Incorrect. In order to provide a framework, you can consider the answer is Correct or Incorrect according to the following criteria:

- The answer is Incorrect if: \{incorrect\}
- The answer is Correct if: \{correct\}

In your discussion try to focus on the NEGATIVE aspect of the answer: why would you consider it as Incorrect? Write your review in English.
\end{PromptVerbatim}
\end{promptbox}

\begin{promptbox}{Metareviewer}
\begin{PromptVerbatim}
You are a teacher in a distant planet. You have received two opinions from colleagues in your solar system about the following question-answer pair in a high-school test:
- Teacher's question: '\{question\}'.
- High-school student's answer: '\{answer\}'.

On one hand, the first colleague says: '\{positive\_review\}'.
On the other hand, the second colleague says: '\{negative\_review\}'.

Your task is to have the final word. Considering that the answer was written by a high-school student, say if it is 'Correct' or 'Incorrect', according to the words of your colleagues and the following criteria:
- The answer is Incorrect if: \{incorrect\}
- The answer is Correct if: \{correct\}

Write a brief essay to support your decision, and then end it saying if it is 'Correct' or 'Incorrect' between two \# (e.g. \#Correct\#).
\end{PromptVerbatim}
\end{promptbox}

\end{document}